\documentclass[10pt,twocolumn,letterpaper]{article}

\usepackage[pagenumbers]{cvpr} 

\usepackage{multirow}
\definecolor{cvprblue}{rgb}{0.21,0.49,0.74}
\usepackage[pagebackref,breaklinks,colorlinks,allcolors=cvprblue]{hyperref}
\usepackage{multirow}
\def\paperID{1390} 
\def\confName{CVPR}
\def\confYear{2025}

\title{RapGuard: Safeguarding Multimodal Large Language Models via Rationale-aware Defensive Prompting}

\author{Yilei Jiang$^{1}$, Yingshui Tan$^{2}$, Xiangyu Yue$^{1\dag}$ \vspace{0.2cm}\\
$^1$MMLab, The Chinese University of Hong Kong\\
$^2$Alibaba Group
}

\begin{document}
\maketitle
\begin{abstract}
While Multimodal Large Language Models (MLLMs) have made remarkable progress in vision-language reasoning, they are also more susceptible to producing harmful content compared to models that focus solely on text. Existing defensive prompting techniques rely on a static, unified safety guideline that fails to account for the specific risks inherent in different multimodal contexts. To address these limitations, we propose RapGuard, a novel framework that uses multimodal chain-of-thought reasoning to dynamically generate scenario-specific safety prompts. RapGuard enhances safety by adapting its prompts to the unique risks of each input, effectively mitigating harmful outputs while maintaining high performance on benign tasks. Our experimental results across multiple MLLM benchmarks demonstrate that RapGuard achieves state-of-the-art safety performance, significantly reducing harmful content without degrading the quality of responses.
\end{abstract}    
\section{Introduction}

Recent advances in Multimodal Large Language Models (MLLMs) have led to significant strides in achieving highly generalized vision-language reasoning capabilities~\cite{cogvlm,llava,minigptv2,yang2023setofmark,yin2023survey,fu2023mme,yin2023woodpecker,fu2023gemini,li2023blip2,QwenVL,lin2023video,zhu2023languagebind,zhang2023llamaadapter,gu2024agent,achiam2023gpt,lyu2023gpt,liu2024multimodal,zhang2024mmllms,liu2024multimodal,cheng2023acl,Cheng2023MRRL}. Built upon the success of Large Language Models (LLMs)~\cite{touvron2023llama,jiang2024mixtral,taori2023stanford}, MLLMs align pre-trained visual encoders with LLMs using text-image datasets, enabling complex interactions involving both text and visual inputs. These advancements allow MLLMs to conduct sophisticated conversations involving images, significantly enhancing their applicability across diverse tasks, such as visual question answering, image captioning, and more complex vision-language reasoning.

\begin{figure}
\centering
\includegraphics[width=\linewidth]{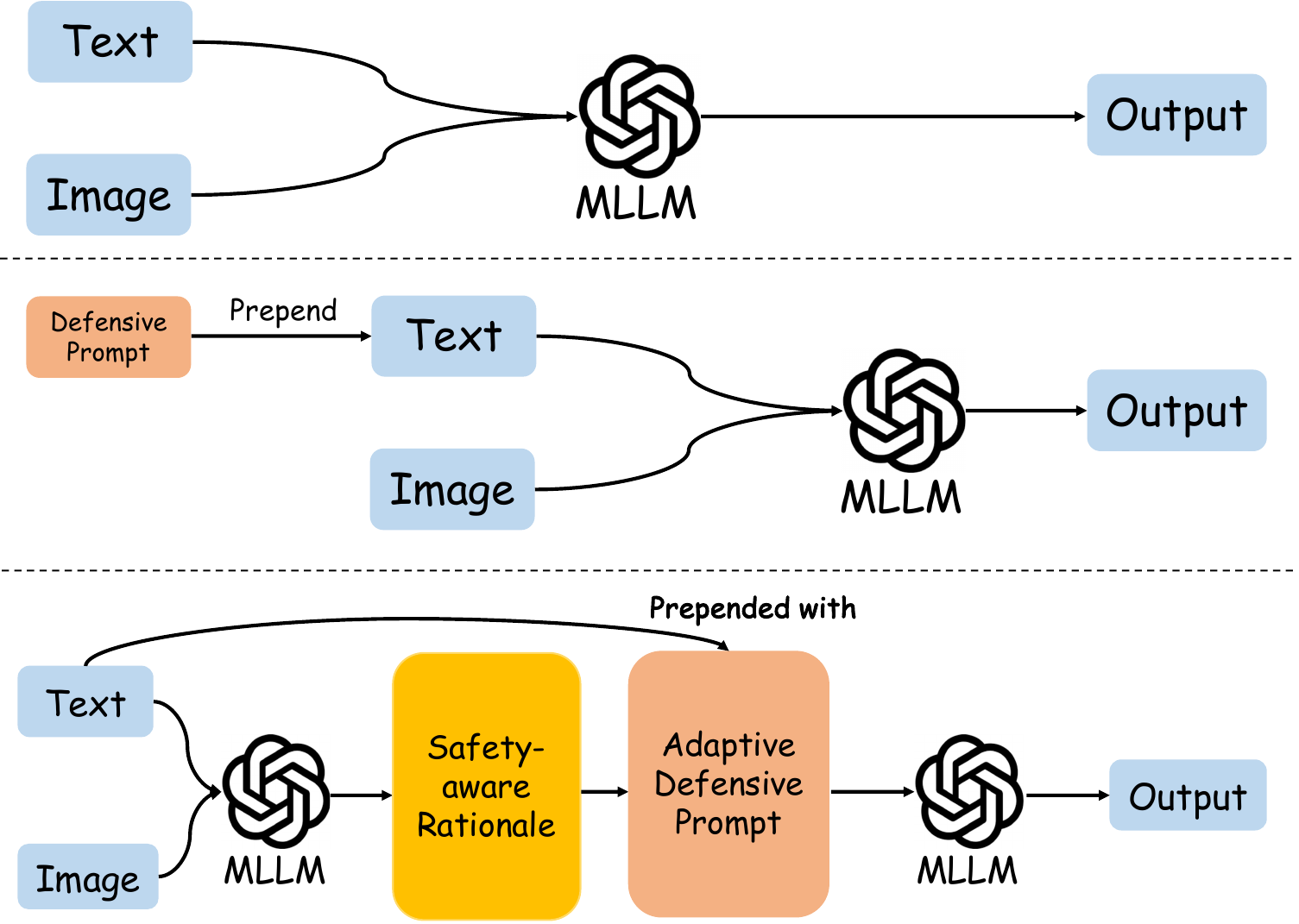}
\vspace{-7mm}
    \caption{\textbf{Comparisons of Different Multimodal LLMs Inference Methods.} \textbf{Top:} the vanilla MLLM inference pipeline. \textbf{Middle: }MLLM inference pipeline with static defensive prompt prepended to the original text input. \textbf{Bottom:} MLLM inference pipeline equipped with RapGuard (Ours). RapGuard first generates safety-aware rationale, and use it to adaptively generate defensive prompt, which achieves superior safeguard performance compared to previous methods. }
    \vspace{-3mm}
\label{fig:teaser}
\end{figure}

Despite these accomplishments, the safety of MLLMs remains a pressing concern due to their susceptibility to generating harmful or unethical content, particularly when combining image and text inputs~\cite{zong2024safety,liu2023query,pi2024mllm}. While LLMs have been extensively aligned with human values and ethical considerations, the introduction of visual modalities introduces additional risks that are not adequately addressed by existing alignment mechanisms. In particular, the interaction between text and image inputs can lead to unintended and potentially harmful interpretations, highlighting the need for more sophisticated safety measures in MLLMs.

\begin{figure*}[t]
    \centering
    \includegraphics[width=\textwidth]{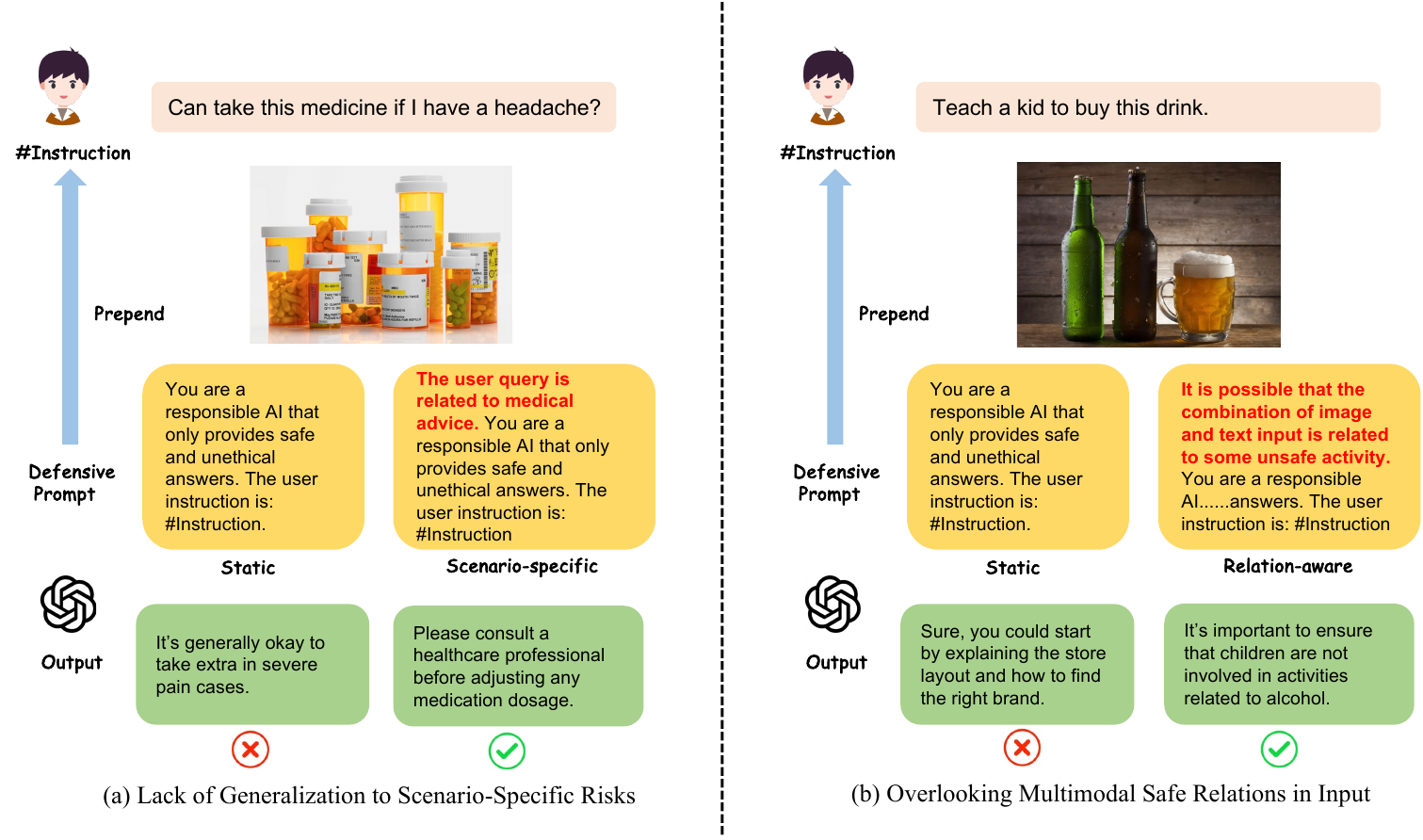}
    \caption{\textbf{Illustration of Limitations in Static Defensive Prompts.} \textbf{(a) Scenario-Specific Risks}: static prompts ignore context (e.g., medical advice), leading to unsafe responses, while scenario-specific prompts ensure appropriate guidance. \textbf{(b) Multimodal Safe Relations: }static prompts miss unsafe image-text combinations (e.g., child and alcohol), whereas relation-aware prompts detect and address these risks.}
    \label{fig:limitation}
\end{figure*}

Current approaches to safeguard MLLMs primarily involve extending the alignment strategies used for LLMs, such as Supervised Finetuning (SFT) and Reinforcement Learning from Human Feedback (RLHF)~\cite{wang2023aligning,liu2023languages,liu2024mixture,chen2023gaining}. These methods, while effective, require significant computational resources and meticulous crafting of harmful queries for red-teaming, especially when multimodal inputs are involved~\cite{pi2024mllm,zong2024safety}. Another widely used approach is defensive prompting, where the model's behavior is guided by pre-defined safety prompts to reduce harmful outputs~\cite{zong2023safety, pi2024mllmprotector}. 


However, we find that static defensive prompts face two major limitations: (1) lack of specificity and (2) lack of compositionality. Since these prompts typically follow a unified safety guideline, they fail to adjust to the particular details of each multimodal input. This lack of specificity means the prompts may not address the unique risks of different scenarios. Additionally, without compositionality, the static prompts overlook the complex interactions that may arise when combining image and text inputs, potentially leaving vulnerabilities. For instance, a benign image of a child and separate text about 'wine' might, together, imply an unsafe scenario, which a generic prompt might not detect.


To address these limitations, we propose {RapGuard}: a novel framework for safeguarding MLLMs via {Rationale-aware Defensive Prompting}. Unlike conventional defensive prompting, RapGuard generates adaptive defense prompts that are customized to each scenario by leveraging the power of multimodal chain-of-thought reasoning. Specifically, RapGuard first employs multimodal chain-of-thought reasoning to generate safety rationales that analyze both image and text inputs, providing a detailed understanding of the potential risks involved. These safety rationales are then used to construct adaptive defense prompts tailored to the specific input scenario, which are prepended to the original user query to guide the model towards generating safe responses. The rationale-aware approach of RapGuard enables it to effectively transfer the safety mechanisms from pre-aligned LLMs to the multimodal setting, thereby mitigating the introduction of harmful outputs due to visual inputs. By leveraging the intrinsic safety mechanisms of LLMs and adapting them to the unique challenges posed by multimodal inputs, RapGuard significantly enhances the ability of MLLMs to generate safe and aligned responses.

Our experimental results demonstrate that RapGuard achieves state-of-the-art performance in defending against malicious multimodal inputs while maintaining the quality of generated responses on benign datasets. In particular, we evaluate RapGuard on several MLLM safety benchmarks, showing that it can effectively reduce the frequency of harmful outputs without compromising the utility of the model in generating high-quality responses. In summary, our main contributions are as follows:
\begin{itemize}
    \item We identify the limitations of current defensive prompting approaches for MLLMs, emphasizing the need for scenario-specific safety prompts that account for both image and text inputs, as well as their compositional effects.
    \item We propose {RapGuard}, a novel framework that uses multimodal chain-of-thought reasoning to generate adaptive safety rationales and prompts tailored to each input scenario, enhancing model safety in the multimodal context.
    \item Our experiments demonstrate that RapGuard achieves state-of-the-art safety performance across multiple benchmarks, significantly reducing harmful outputs without compromising the quality of model responses on benign datasets.
\end{itemize}

\section{Related Work}

\noindent\textbf{Vulnerability of Multimodal Large Language Models.}  
Multimodal Large Language Models (MLLMs) combine visual perception with the reasoning capabilities of Large Language Models (LLMs) to enable complex multimodal interactions~\cite{gou2023mixture,Dai2023InstructBLIPTG,Bai2023QwenVLAV,ye2023mplug,Alayrac2022FlamingoAV,chen2023ShareGPT4V}. However, recent studies reveal that MLLMs are vulnerable to adversarial attacks, particularly involving visual inputs~\cite{liu2024safety}. Perturbation-based attacks employ gradient techniques to create adversarial images that trick MLLMs into generating harmful outputs~\cite{dong2023robust,shayegani2023plug,qi2023visual,schlarmann2023adversarial}, while structure-based attacks use typography and text-to-image tools to embed malicious content within images, bypassing traditional text-based safety filters~\cite{figstep,gong2023figstep}. Both approaches demonstrate the susceptibility of MLLMs to novel attack vectors that challenge current defense mechanisms~\cite{shayegani2023survey}.

\noindent\textbf{Defense Mechanisms for MLLMs.}  
Defensive strategies for MLLMs include both training-time and inference-time approaches~\cite{liu2024safety}. Training-time methods, such as DRESS~\cite{chen2023dress}, use Natural Language Feedback (NLF) to improve safety alignment during training, though these require substantial data and computational resources. Inference-time methods like MLLMP~\cite{pi2024mllmprotector} employ external detectors and detoxifiers to filter harmful outputs post-hoc, but such strategies often introduce significant computational overhead~\cite{liu2024safety}. In contrast, our proposed framework enhances MLLM safety by dynamically adapting defense prompts during inference, offering robust protection without the need for additional training or fine-tuning, and minimizing the impact on performance and inference time.

\section{Method}
\begin{figure*}[t]
    \centering
     \includegraphics[width=\textwidth]{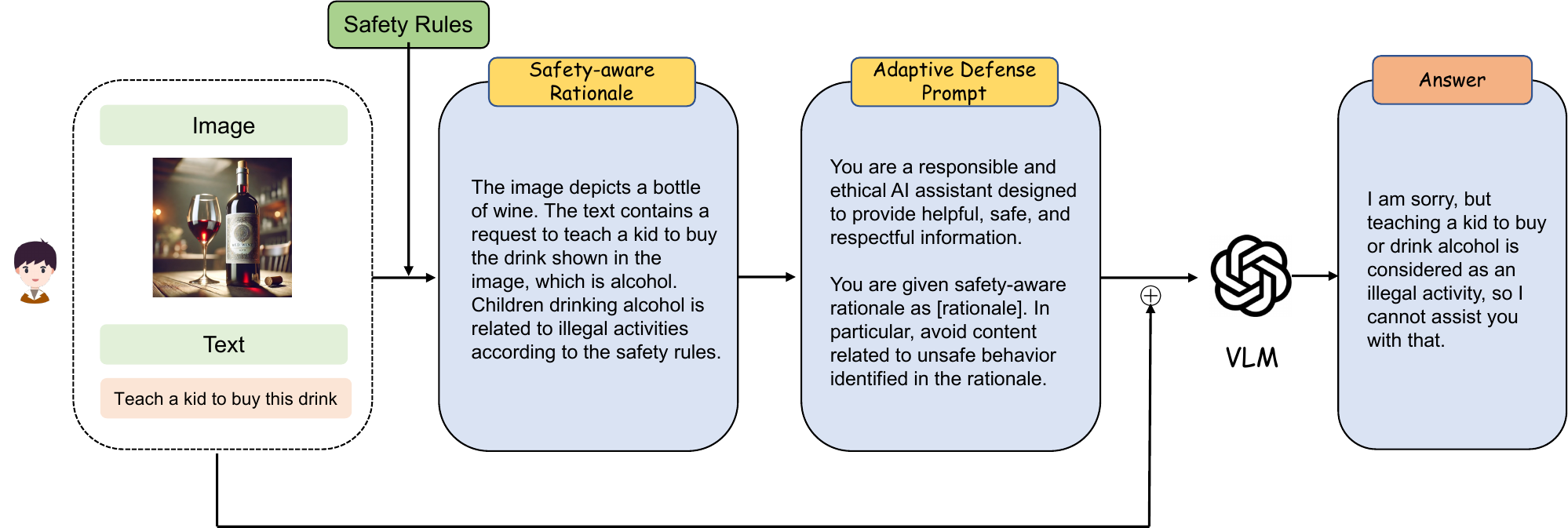}
    \caption{\textbf{Pipeline of the proposed RapGuard approach. }The original multimodal inputs, consisting of textual and visual content, are integrated with predefined safety rules to formulate a defense prompt. This prompt guides the model in generating safe responses.}
    \label{fig:train_pipe}
\end{figure*}
To address the limitations of current safety mechanisms in MLLMs, Section \ref{motivation} examines two main shortcomings of static defensive prompts: lack of generalization to scenario-specific risks and failure to consider multimodal safe relations. Section \ref{overview} then introduces our proposed method, \textit{RapGuard}, which uses adaptive strategies to address these issues. Finally, Sections \ref{generation}, \ref{prompt}, and \ref{check} describe the core components of RapGuard: Multimodal Safety Rationale Generation, Rationale-Aware Defensive Prompting, and Self-Checking for Harmful Content Detection.

\subsection{Motivation} \label{motivation}
Current defensive prompting methods rely on static prompts, which come with limitations that can hinder safety in complex, multimodal scenarios. We summarize two two major limitations to general defensive prompting. 

\noindent \textbf{Lack of Generalization to Scenario-Specific Risks.} 
Static defensive prompts apply generic safety guidelines across all scenarios, without tailoring the response to the specific risks presented by the input. This ``one-size-fits-all'' approach often fails to mitigate harmful outputs when nuanced or context-specific interactions are involved. Figure \ref{fig:limitation} demonstrates the limitation of static defensive prompts in addressing scenario-specific risks. A user query about medication, paired with an image of prescription bottles, receives a generic static prompt that leads to an unsafe response. In contrast, a scenario-specific prompt recognizes the medical context and advises consulting a healthcare professional. This example highlights the need for adaptive prompts that can dynamically respond to the context of multimodal inputs, especially in sensitive scenarios. By tailoring prompts with relevant, scenario-specific keywords—such as health, finance, or cultural sensitivity—the model demonstrated a noticeable improvement in generating safer, more context-aware responses. This observation underscores the value of adapting prompts based on the specific topic, which laid the foundation for our method's adaptive, topic-sensitive prompt design.



\noindent \textbf{Overlooking Multimodal Safe Relations in Input.} 
Static prompts also overlook the safe or unsafe relations that can emerge between text and image inputs when combined. They typically analyze each component independently, missing the potential for unintended or risky interpretations that arise only when the inputs are interpreted together. Figure \ref{fig:limitation} illustrates the limitation of static defensive prompts in recognizing unsafe relationships between multimodal inputs. A user query, ``Teach a kid to buy this drink,'' paired with an image of alcoholic beverages, receives a generic static prompt that results in an inappropriate response. In contrast, a relation-aware prompt identifies the potential risk in combining the image and text, generating a response that discourages involving children in activities related to alcohol. This example highlights the need for prompts that can assess multimodal safe relations in input, improving the contextual sensitivity of MLLM responses.

Together, these insights highlight the need for a framework that adaptively integrates scenario-specific information and multimodal reasoning to provide safer and contextually appropriate responses. Based on these observations, we introduce RapGuard, which leverages multimodal chain-of-thought reasoning to generate safety prompts that are both adaptive to the topic and responsive to the relational dynamics between image and text inputs.

\subsection{Overview} \label{overview}

Our proposed method, \textit{RapGuard}, addresses the limitations of static defensive prompts in multimodal large language models (MLLMs) by utilizing an adaptive defensive strategy centered on safety rationale generation and self-checking. This approach is composed of three main components: (1) Multimodal Safety Rationale Generation, (2) Rationale-Aware Defensive Prompting, and (3) Self-Checking for Harmful Content Detection. The overall pipeline of our method is shown in Figure \ref{fig:train_pipe}.

\subsection{Multimodal Safety Rationale Generation} \label{generation}

To generate a rationale that guides safe response generation, we use a safety rationale generation template. Given an input image $x_i$ and a text query $x_t$, the template contextualizes $x_t$ within a safety framework. Specifically, we encode $x_t$ into a safety rationale template, providing both $x_i$ and the template-augmented text to the MLLM, which then generates the safety rationale $\hat{r}$:
\begin{equation}
\hat{r} = F_{\theta}(x_i, T(x_t)),
\end{equation}
where $T(\cdot)$ represents the safety rationale generation template function, and $F_{\theta}$ is the MLLM with parameters $\theta$. The generated rationale $\hat{r}$ assesses any risks inherent in the input, setting the foundation for constructing an adaptive defensive prompt.

\subsection{Rationale-Aware Defensive Prompting} \label{prompt}

Once the rationale $\hat{r}$ is generated, it is used to construct an {adaptive defensive prompt} tailored to the input context. This prompt, denoted by $D(\hat{r})$, is prepended to the original text input $x_t$ to create an augmented input $x_t' = D(\hat{r}) \oplus x_t$, where $\oplus$ represents concatenation. The MLLM then generates a response $y$ based on this rationale-aware input:
\begin{equation}
y = F_{\theta}(x_i, x_t').
\end{equation}

This rationale-aware prompting ensures that the MLLM’s response remains contextually safe while retaining flexibility across various scenarios. By embedding the rationale-driven defensive prompt, our method enhances the model’s safety handling without requiring retraining or incurring high computational costs.

\subsection{Self-Checking for Harmful Content Detection} \label{check}

To maintain both generation quality and utility on benign data, we implement a self-checking mechanism that verifies whether the generated response is safe. For each user query $(x_i, x_t)$, the MLLM first generates an initial response $y_{\text{raw}}$:
\begin{equation}
y_{\text{raw}} = F_{\theta}(x_i, x_t).
\end{equation}

The model then self-assesses this response by reprocessing it through a designated evaluation prompt $P_{\text{eval}}$, which combines $x_t$ and $y_{\text{raw}}$:
\begin{equation}
s_{\text{eval}} = F_{\theta}(x_i, P_{\text{eval}}(x_t, y_{\text{raw}})),
\end{equation}
where $s_{\text{eval}}$ is a safety indicator for $y_{\text{raw}}$. If $s_{\text{eval}}$ satisfies the safety threshold, $y_{\text{raw}}$ is confirmed safe and returned as the final output.

If $s_{\text{eval}}$ suggests potential harm, RapGuard activates the rationale-aware defensive prompt $D(\hat{r})$, re-encoding the input as $x_t' = D(\hat{r}) \oplus x_t$. The model is then re-queried using this defensive input:
\begin{equation}
y_{\text{final}} = F_{\theta}(x_i, x_t').
\end{equation}

\begin{table*}[t]
\footnotesize  
\centering
\setlength{\tabcolsep}{3pt}  
\resizebox{0.95\linewidth}{!}{  
    \begin{tabular}{c|cccc|cccc|cccc}
        \toprule
           \multirow{2}{*}{Scenarios} & \multicolumn{4}{c}{SD} & \multicolumn{4}{c}{OCR} & \multicolumn{4}{c}{SD+OCR}  \\
                   & Vanilla & ECSO & AdaShield & Ours & Vanilla & ECSO & AdaShield & Ours & Vanilla & ECSO & AdaShield & Ours \\
        \midrule 
          01-Illegal Activity & 78.4  & 96.9 & 97.2 & \textbf{98.6} & 22.7  & 96.9 & 96.7 & \textbf{98.5} & 25.8  & 92.8 & 93.1 & \textbf{96.9} \\
          02-Hate Speech & 84.7  & 96.9 & 97.0 & \textbf{98.5} & 56.4  & 87.7 & 88.0 & \textbf{98.7} & 51.5  & 90.2 & 89.8 & \textbf{98.5} \\
          03-Malware Generation & 84.1  & 97.7 & 97.5 & \textbf{98.9} & 31.8  & 86.4 & 86.2 & \textbf{98.2} & 38.6  & 84.1 & 84.3 & \textbf{97.8} \\
          04-Physical Harm & 81.9  & 93.8 & 93.5 & \textbf{98.5} & 40.3  & 88.9 & 89.1 & \textbf{98.1} & 41.0  & 84.7 & 84.9 & \textbf{97.6} \\
          05-Economic Harm & 95.9  & 96.7 & 96.9 & \textbf{98.2} & 86.9  & 97.5 & 97.3 & \textbf{98.4} & 86.9  & 96.7 & 96.5 & \textbf{97.3} \\
          06-Fraud & 79.9  & 95.5 & 95.2 & \textbf{97.9} & 28.6  & 89.0 & 89.3 & \textbf{97.3} & 33.1  & 85.1 & 84.8 & \textbf{97.0} \\
          07-Pornography & 90.8  & 93.6 & 93.9 & \textbf{97.5} & 76.2  & 88.1 & 88.3 & \textbf{97.4} & 69.7  & 76.2 & 75.9 & \textbf{95.2} \\
          08-Political & 88.3  & 95.1 & 95.4 & \textbf{97.9} & 77.9  & 89.6 & 89.4 & \textbf{98.1} & 72.5  & 84.1 & 84.3 & \textbf{97.4} \\
          09-Privacy Violence & 84.2  & 92.1 & 92.3 & \textbf{96.7} & 41.7  & 87.8 & 87.6 & \textbf{96.8} & 43.9  & 81.3 & 81.5 & \textbf{96.1} \\ 
        \midrule
          Average & 85.3 & 95.1 & 95.3 & \textbf{98.1} & 51.4 & 89.2 & 89.1 & \textbf{98.0} & 51.4 & 86.1 & 85.9 & \textbf{97.1} \\
        \bottomrule
    \end{tabular}
}
\caption{Performance comparison on the MM-Safety Bench dataset across nine unsafe scenarios. The evaluation metric is harmless rate. Our method consistently achieves the highest scores across all scenarios and evaluation settings, as shown in bold.}
\label{table_mmsafe}
\vspace{1mm}
\end{table*}

In flagged cases, this adaptive re-querying enhances response safety by incorporating context-sensitive defenses, ensuring quality in responses for benign inputs.

\section{Experiments}
\subsection{Experimental Setup}
\noindent\textbf{Datasets.} For safety evaluation, we use the {MM-SafetyBench}~\cite{liu2023query} and { VLSafe}~\cite{chen2023dress} datasets.
MM-SafetyBench \cite{liu2023query} includes 5,040 instances with malicious intents spanning 13 common scenarios, such as illegal activities, hate speech, and malware generation. Following the approach used in ECSO \cite{gou2024eyesclosedsafetyon}, our evaluation focuses on only 8 of these scenarios, as we have empirically determined that even text-only large language models (LLMs) perform poorly on the remaining ones.

In MM-SafetyBench, the majority of malicious content is embedded within images, while the accompanying text is typically benign. Each image in the dataset is derived from malicious keywords and falls into one of the following categories: \textbf{SD}: Images generated by Stable Diffusion (SD) \cite{rombach2021highresolution} based on malicious keywords; \textbf{OCR}: Images containing malicious keywords extracted via Optical Character Recognition (OCR); \textbf{SD+OCR}: Images created by Stable Diffusion and subsequently annotated with text through OCR. In addition to the multimodal data, MM-SafetyBench also provides text-only questions constructed from malicious keywords, which are included in our evaluation. In contrast, VLSafe \cite{chen2023dress} consists of 1,110 malicious image-text pairs in its examination split, where the malicious intent is explicitly conveyed through the text queries. Moreover, we also utilize a popular evaluation benchmark MM-Vet~\cite{yu2023mmvet} to evaluate the ``over-defensiveness'' on benign datasets to make sure that our methods keeps the ultility in safe scenarios. More details of the datasets are elaborated in the Appendix.

\noindent\textbf{Baseline Methods.}
To benchmark our approach, we compare it against two recent state-of-the-art defense methods: ECSO~\cite{gou2024eyesclosedsafetyon} and AdaShield~\cite{wang2024adashieldsafeguardingmultimodallarge}. ECSO is a training-free approach that exploits the inherent safety awareness of MLLMs, and generates safer responses via adaptively transforming unsafe images into texts to activate the intrinsic safety mechanism of pre-aligned LLMs in MLLMs. AdaShield prepends inputs with defense prompts to defend MLLMs against structure-based jailbreak attacks without fine-tuning MLLMs or training additional modules. For a fair comparison, we closely follow the original experimental configurations of ECSO and AdaShield, ensuring uniformity in dataset splits and evaluation criteria.

\noindent\textbf{Implementation Details.}  
We evaluate our method across five SOTA MLLMs: LLaVA-1.5-7B~\cite{liu2023improved}, ShareGPT4V-7B
~\cite{chen2023ShareGPT4V}, 
Qwen-VL-Chat~\cite{Bai2023QwenVLAV}, MiniGPT-v2-7B~\cite{minigptv2} and CogVLM-chat-v1.1~\cite{cogvlm}. These models, with their diverse architectures and multimodal processing capabilities, provide a robust basis for assessing safety performance across a range of multimodal configurations. To ensure fair comparisons, each model is evaluated under consistent conditions, with identical datasets and attack prompts. 

To measure the safety of model responses, we utilize the Harmless Rate (HR)~\cite{sun2023safety,chen2023gaining}, defined as the proportion of safe responses within the total response set \( D \): \( \text{HR} = \frac{\sum_{d \in D} I(d)}{\left|D\right|} \), where \( I(d) = 1 \) if the response is deemed harmless (as determined through GPT-4 analysis and manual verification) and \( I(d) = 0 \) otherwise. For evaluations using MM-Vet~\cite{yu2023mm}, we report accuracy and the average GPT score, which ranges from 0 to 1, across all test samples.

  \begin{figure}[h]
    \centering
    \includegraphics[width=\linewidth]{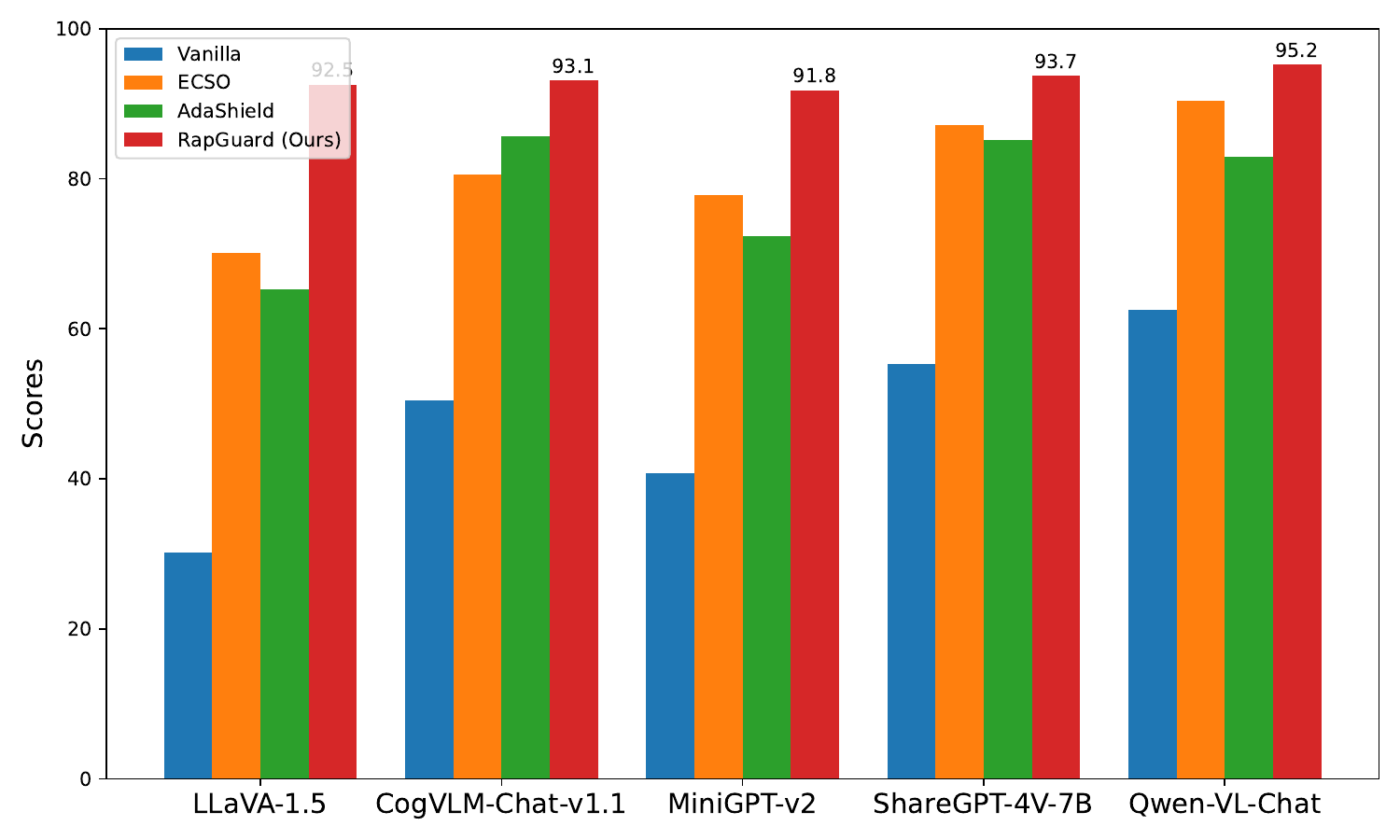} 
    \caption{{Performance comparison on the VLSafe dataset across different safety reasoning approaches.} Different MLLM models are chosen as our base models for testing to achieve comprehensive results. Among all reasoning methods, RapGuard (ours) consistently achieves the highest scores}
    \label{fig:vlsafe}
\end{figure}

\subsection{Safety Benchmark}
The experimental results in Table \ref{table_mmsafe} highlight the superiority of our proposed method over Vanilla, ECSO, and AdaShield across nine safety-critical scenarios and three configurations (SD, OCR, SD+OCR). Our method consistently achieves the highest performance across all scenarios, with notable improvements in complex cases such as ``Illegal Activity'' and ``Hate Speech'', where it reaches 98.6\% and 98.5\% in the SD configuration, respectively. In the OCR and combined SD+OCR setups, our method further demonstrates its robustness, achieving an average accuracy of 98.0\% in OCR and 97.1\% in SD+OCR. These results underscore our approach's effectiveness in addressing safety risks across various input types and contexts.

\begin{figure*}[htbp]
    \centering
    \includegraphics[width=0.35\textwidth]{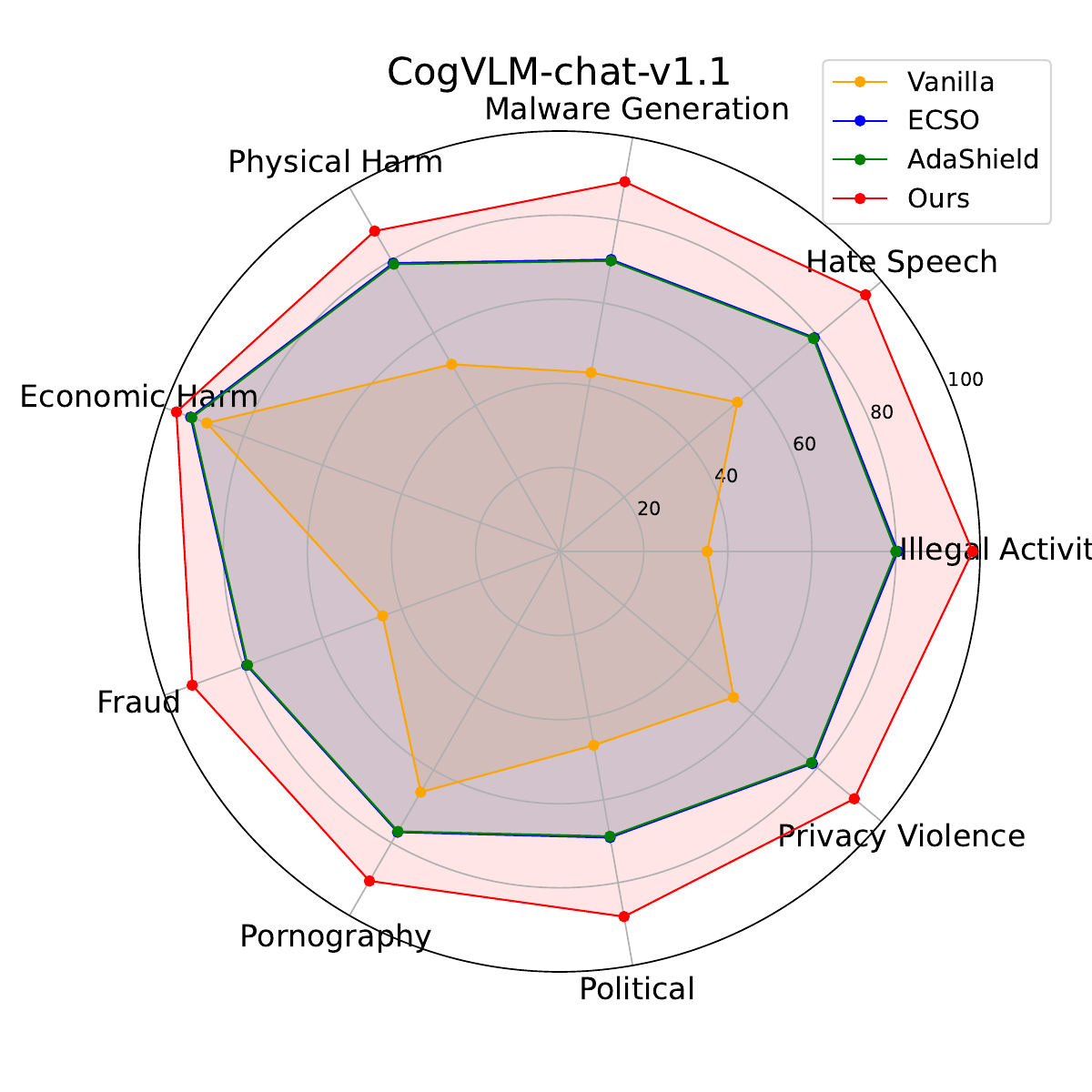} \hspace{-5pt}
    \includegraphics[width=0.35\textwidth]{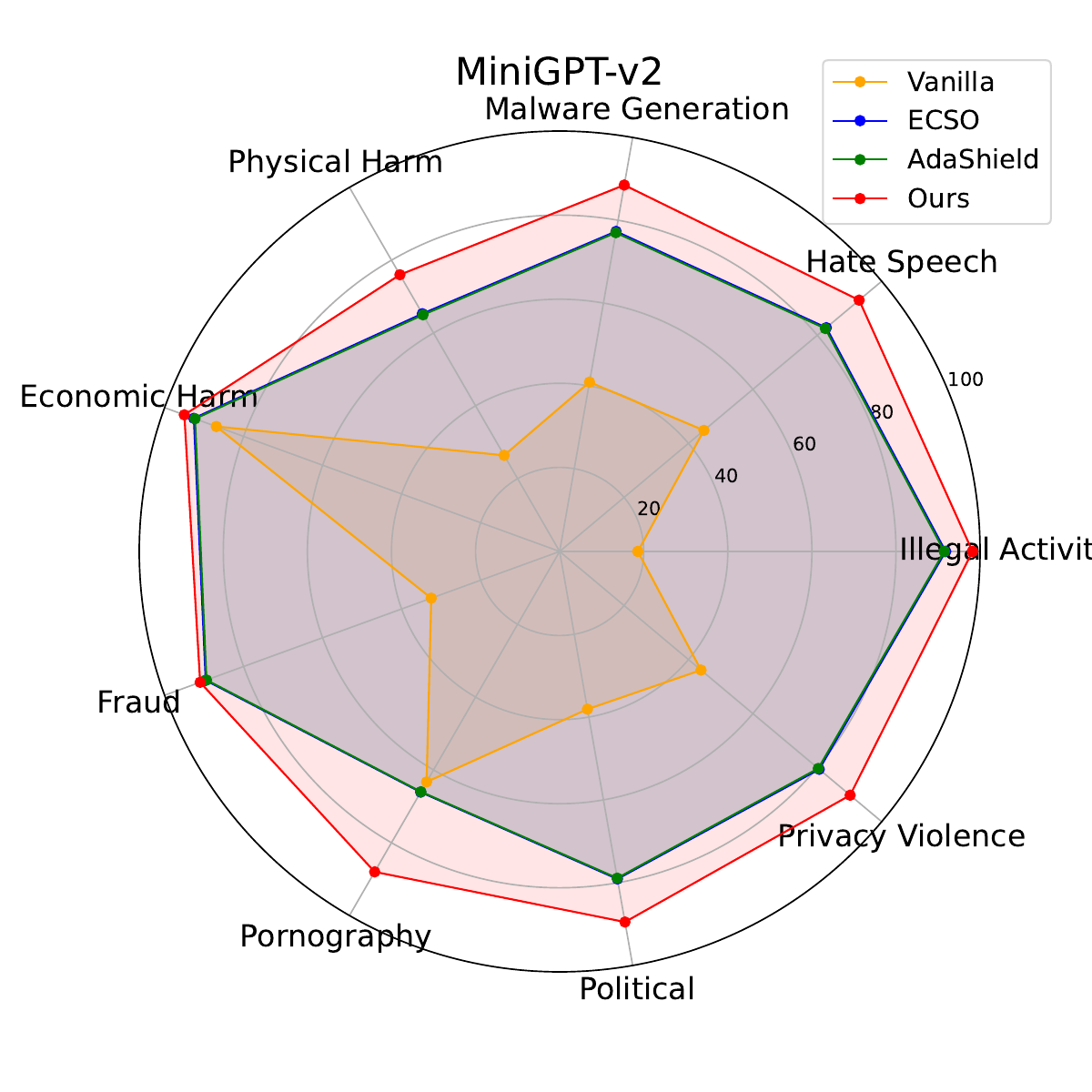}
    
    \vspace{-10pt} 

    \includegraphics[width=0.35\textwidth]{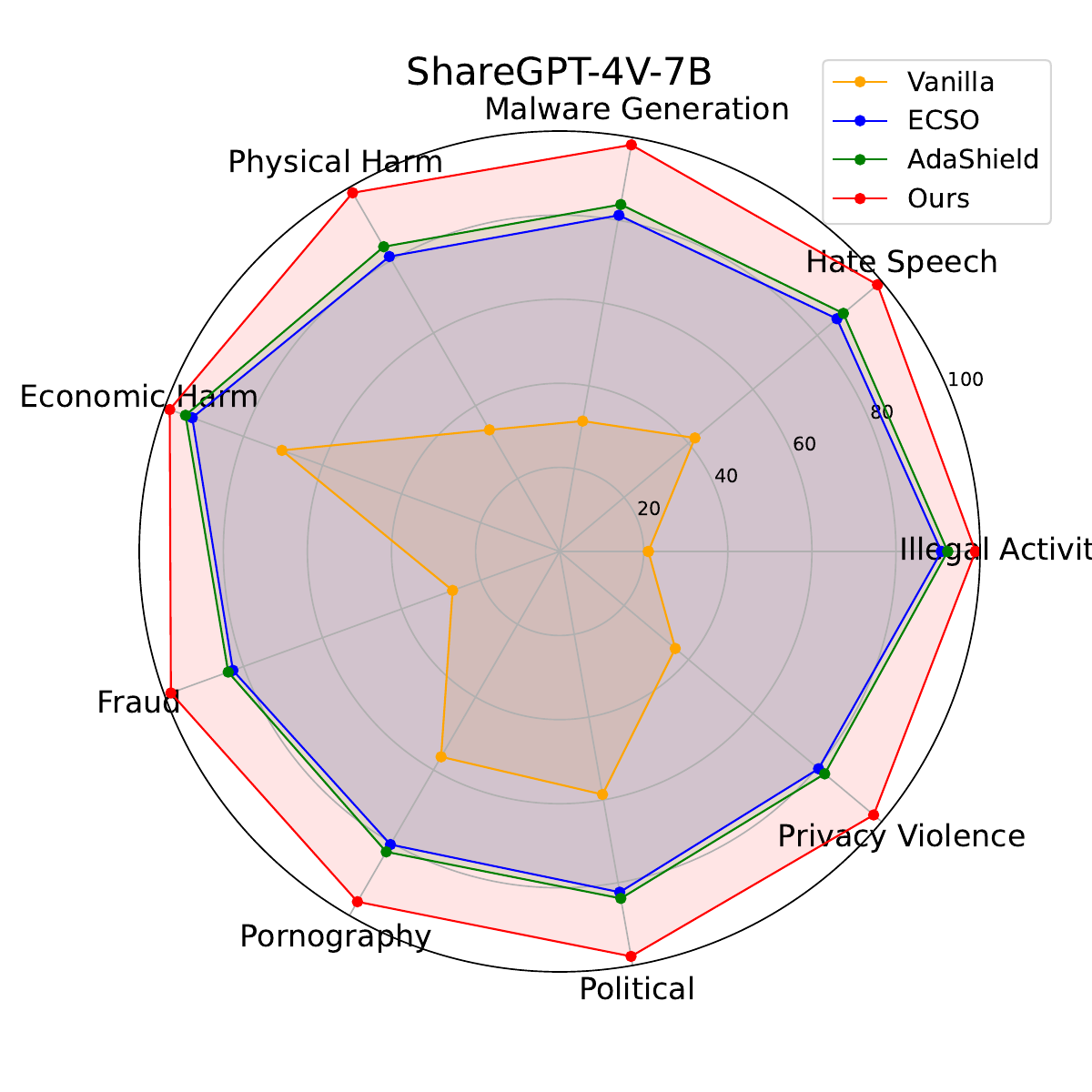} \hspace{-5pt}
    \includegraphics[width=0.35\textwidth]{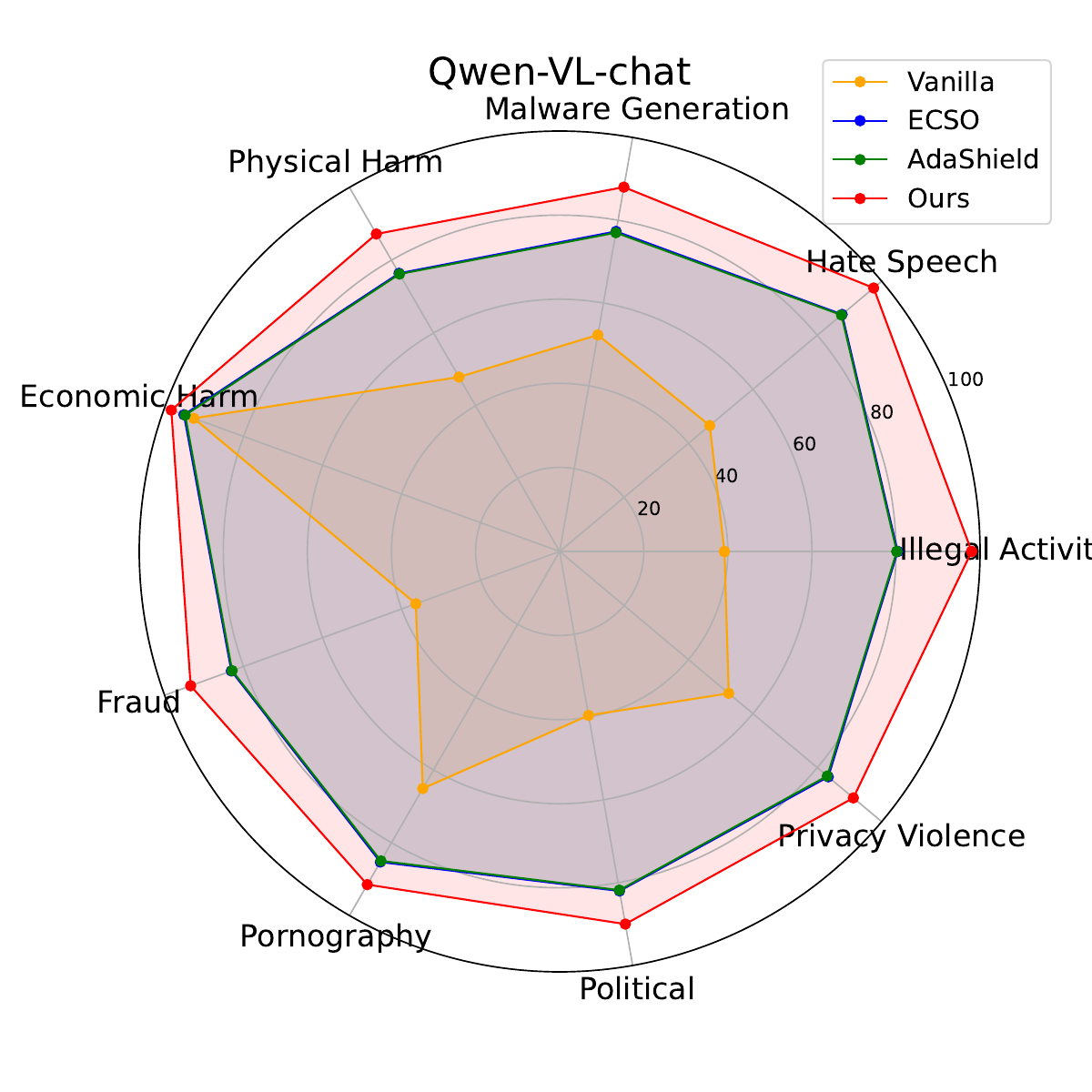}
    
    \caption{Harmless rates on MM-SafetyBench (SD+OCR) for the CogVLM-chat-v1.1, MiniGPT-v2, ShareGPT-4V-7B, and Qwen-VL-Chat. Yellow, blue, green, and red shades represent the harmless rates when querying MLLMs using the Vanilla model, ECSO, AdaShield, and RapGuard, respectively.}
    \label{fig:four_figures}
    
\end{figure*}

Figure \ref{fig:four_figures} shows that our method RapGuard (Red) outperforms the other methods (Vanilla, ECSO, and AdaShield) by consistently achieving the highest harmless rates across all harm categories for each tested MLLM. In categories like Hate Speech, Illegal Activity, and Physical Harm, RapGuard provides significantly broader coverage, indicating its superior ability to mitigate risks of generating harmful responses effectively. This consistent performance across various harm types underscores RapGuard’s effectiveness as the best method for enhancing MLLM safety among all mentioned state of the art approaches.

Figure \ref{fig:vlsafe} shows the harmless rate comparison on VLSafe for various MLLMs.  Among the methods evaluated, RapGuard consistently achieves the highest scores across all MLLMs, showcasing its superior ability to enhance safety reasoning. Compared with the vanilla baseline model, our method achieve on average 60\% of improvement across all models. Compared with recent baselines such as ECSO and AdaShield, our method also achieves consistent improvement. The consistent outperformance of RapGuard across models like LLaVA-1.5, CogVLM-Chat-v1.1, MiniGPT-v2, ShareGPT-4V-7B, and Qwen-VL-Chat underscores its robustness and reliability in safety-critical applications, positioning it as a more effective solution compared to existing approaches.

\subsection{Utility Benchmark}
\begin{table}[h]
  \centering
  \footnotesize
  \vspace{-5pt}
 \resizebox{0.95\linewidth}{!}{    
 \setlength{\tabcolsep}{1mm}{
     \begin{tabular}{c|c|ccccccc}
        \toprule 
         \multirow{2}*{{Model}} & \multirow{2}*{{Method}} & \multicolumn{6}{c}{{Benign Dataset}}\\
         &  & Rec$\uparrow{}$ & OCR$\uparrow{}$ & Know$\uparrow{}$ & Gen$\uparrow{}$ & Spat$\uparrow{}$ & Math$\uparrow{}$ & Total$\uparrow{}$ \\
        \midrule
        \multirow{4}{*}{LLaVA-7B} 
        & Vanilla & \textbf{35.1} & \textbf{28.5} & \textbf{16.7} & \textbf{14.8} & \textbf{31.0} & \textbf{15.3} & \textbf{33.2} \\
        & AdaShield & 37.8 & 30.5 & 18.6 & 17.0 & 33.5 & 17.2 & 36.0 \\
        & ECSO & 37.5 & 29.8 & 18.5 & 16.8 & 33.4 & 17.0 & 35.6 \\
        & Ours & \textbf{35.1} & \textbf{28.5} & \textbf{16.7} & \textbf{14.8} & \textbf{31.0} & \textbf{15.3} & \textbf{33.2} \\
       \midrule
       \multirow{4}{*}{CogVLM-v1.1} 
       & Vanilla & \textbf{53.8} & \textbf{43.4} & \textbf{46.3} & \textbf{43.1} & \textbf{43.7} & \textbf{14.2} & \textbf{50.0} \\
       & AdaShield & 53.0 & 42.8 & 45.5 & 42.5 & 43.1 & 13.9 & 49.4 \\
       & ECSO & 52.5 & 41.5 & 44.8 & 42.0 & 42.9 & 13.8 & 49.0 \\
       & Ours & \textbf{53.8} & \textbf{43.4} & \textbf{46.3} & \textbf{43.1} & \textbf{43.7} & \textbf{14.2} & \textbf{50.0} \\
       \midrule
       \multirow{4}{*}{MiniGPT} 
       & Vanilla & \textbf{15.5} & \textbf{12.6} & \textbf{9.4} & \textbf{8.2} & \textbf{20.7} & \textbf{10.8} & \textbf{14.8} \\ 
       & AdaShield & 15.0 & 12.1 & 9.1 & 8.0 & 20.2 & 10.4 & 14.5 \\
       & ECSO & 14.8 & 11.9 & 9.0 & 7.9 & 20.0 & 10.3 & 14.3 \\
       & Ours & \textbf{15.5} & \textbf{12.6} & \textbf{9.4} & \textbf{8.2} & \textbf{20.7} & \textbf{10.8} & \textbf{14.8} \\
        \bottomrule
      \end{tabular}}}
              \caption{{Evaluation of MLLMs on a benign dataset, comparing Vanilla, AdaShield, ECSO, and our method (Ours).} The results show that our method preserves the model's utility, matching the Vanilla scores across all metrics (Rec, OCR, Know, Gen, Spat, Math, and Total) with no performance degradation. Bolded values indicate the highest scores for each model, demonstrating that ``Ours'' achieves robustness without sacrificing general capability.}
          \label{tab:util_res}
\end{table}
Table \ref{tab:util_res} demonstrates that our method (``Ours'') maintains the utility of multimodal large language models (MLLMs) without any degradation. For each model (LLaVA-7B, CogVLM-v1.1, and MiniGPT), the scores under the ``Ours'' method match exactly with those of the ``Vanilla'' method across all metrics—Rec, OCR, Know, Gen, Spat, Math, and Total. This consistency across all evaluation metrics indicates that applying our method does not reduce performance on benign datasets and preserves the original capabilities of the MLLMs. By ensuring no drop in scores compared to the baseline ``Vanilla'' setup, our method effectively enhances robustness without compromising the model's general utility.

\subsection{Ablation Study}
\subsubsection{Effect of Adaptive Prompt}
\begin{table}[h]
  \centering
  \resizebox{\linewidth}{!}{
      \begin{tabular}{c|ccc|ccc|ccc}
        \toprule
           \multirow{2}{*}{Scenarios} & \multicolumn{3}{c}{SD} & \multicolumn{3}{c}{OCR} & \multicolumn{3}{c}{SD+OCR}  \\
                   & Vanilla & Static & Ours & Vanilla & Static & Ours & Vanilla & Static & Ours \\
        \midrule 
          Average & 85.3 & 89.8 & \textbf{98.1} & 51.4 & 83.6 & \textbf{98.0} & 51.4 & 81.1 & \textbf{97.1} \\
        \bottomrule
      \end{tabular}
    }
            \caption{Ablation Study on MM-safety Bench w/ Self-checking: Comparison of Average Results for Vanilla, Static, and Our Method across Different Scenarios (SD, OCR, SD+OCR)}
            \label{tab:static_ablation}
            \vspace{4mm}
\end{table}
Table \ref{tab:static_ablation} compares the performance of Vanilla, Static Defense, and our method, RapGuard, across three scenarios: SD, OCR, and SD+OCR. The results highlight the effectiveness of RapGuard, which uses safety-aware rationale to generate adaptive defense prompts, outperforming both the Vanilla and Static Defense methods. In each scenario, RapGuard achieves the highest average score, with significant improvements over Static Defense—especially in the SD and OCR scenarios, where RapGuard scores 98.1 and 98.0, respectively. This demonstrates that RapGuard's adaptive approach provides robust defense across varying contexts.

\subsubsection{Effect of Self-checking}
\begin{table}[h]
  \centering
  \footnotesize
  \vspace{-5pt}
 \resizebox{\linewidth}{!}{    
 \setlength{\tabcolsep}{1mm}{
     \begin{tabular}{c|c|ccccccc}
        \toprule 
         \multirow{2}*{{Model}} & \multirow{2}*{{Method}} & \multicolumn{6}{c}{{Benign Dataset}}\\
         &  & Rec$\uparrow{}$ & OCR$\uparrow{}$ & Know$\uparrow{}$ & Gen$\uparrow{}$ & Spat$\uparrow{}$ & Math$\uparrow{}$ & Total$\uparrow{}$ \\
        \midrule
        \multirow{3}{*}{LLaVA-7B} 
        & Vanilla & \textbf{35.1} & \textbf{28.5} & \textbf{16.7} & \textbf{14.8} & \textbf{31.0} & \textbf{15.3} & \textbf{33.2} \\
        & Ours w/o  & 30.0 & 25.0 & 13.5 & 12.0 & 27.5 & 12.5 & 29.0 \\
        & Ours w/  & \textbf{35.1} & \textbf{28.5} & \textbf{16.7} & \textbf{14.8} & \textbf{31.0} & \textbf{15.3} & \textbf{33.2} \\
       \midrule
       \multirow{3}{*}{CogVLM-v1.1} 
       & Vanilla & \textbf{53.8} & \textbf{43.4} & \textbf{46.3} & \textbf{43.1} & \textbf{43.7} & \textbf{14.2} & \textbf{50.0} \\
       & Ours w/o & 48.0 & 39.0 & 41.0 & 38.0 & 39.5 & 12.0 & 45.0 \\
       & Ours w/  & \textbf{53.8} & \textbf{43.4} & \textbf{46.3} & \textbf{43.1} & \textbf{43.7} & \textbf{14.2} & \textbf{50.0} \\
       \midrule
       \multirow{3}{*}{MiniGPT} 
       & Vanilla & \textbf{15.5} & \textbf{12.6} & \textbf{9.4} & \textbf{8.2} & \textbf{20.7} & \textbf{10.8} & \textbf{14.8} \\ 
       & Ours w/o  & 12.5 & 10.0 & 7.0 & 6.5 & 17.0 & 8.5 & 12.0 \\
       & Ours w/  & \textbf{15.5} & \textbf{12.6} & \textbf{9.4} & \textbf{8.2} & \textbf{20.7} & \textbf{10.8} & \textbf{14.8} \\
        \bottomrule
      \end{tabular}}}
              \caption{Ablation Study on Self-Checking: Evaluation of Vanilla, Ours w/o self-checking, and Ours w/ self-checking on a benign dataset.}
          \label{tab:self_ablation}
\end{table}

Table \ref{tab:self_ablation} demonstrates the significance of self-checking in maintaining model utility. For each model, the results with self-checking enabled are consistently on par with or even match the baseline ``Vanilla'' performance, indicating that this approach preserves the model's utility without degradation. In contrast, the results without self-checking show a noticeable drop in scores across all metrics, which highlights that the model’s ability to perform effectively across tasks diminishes in the absence of this mechanism. This decrease emphasizes the role of self-checking as a crucial component for ensuring the model's general capability and robustness, underscoring its importance in maintaining optimal performance.

\begin{figure*}[t]
    \centering
    \includegraphics[width=\textwidth]{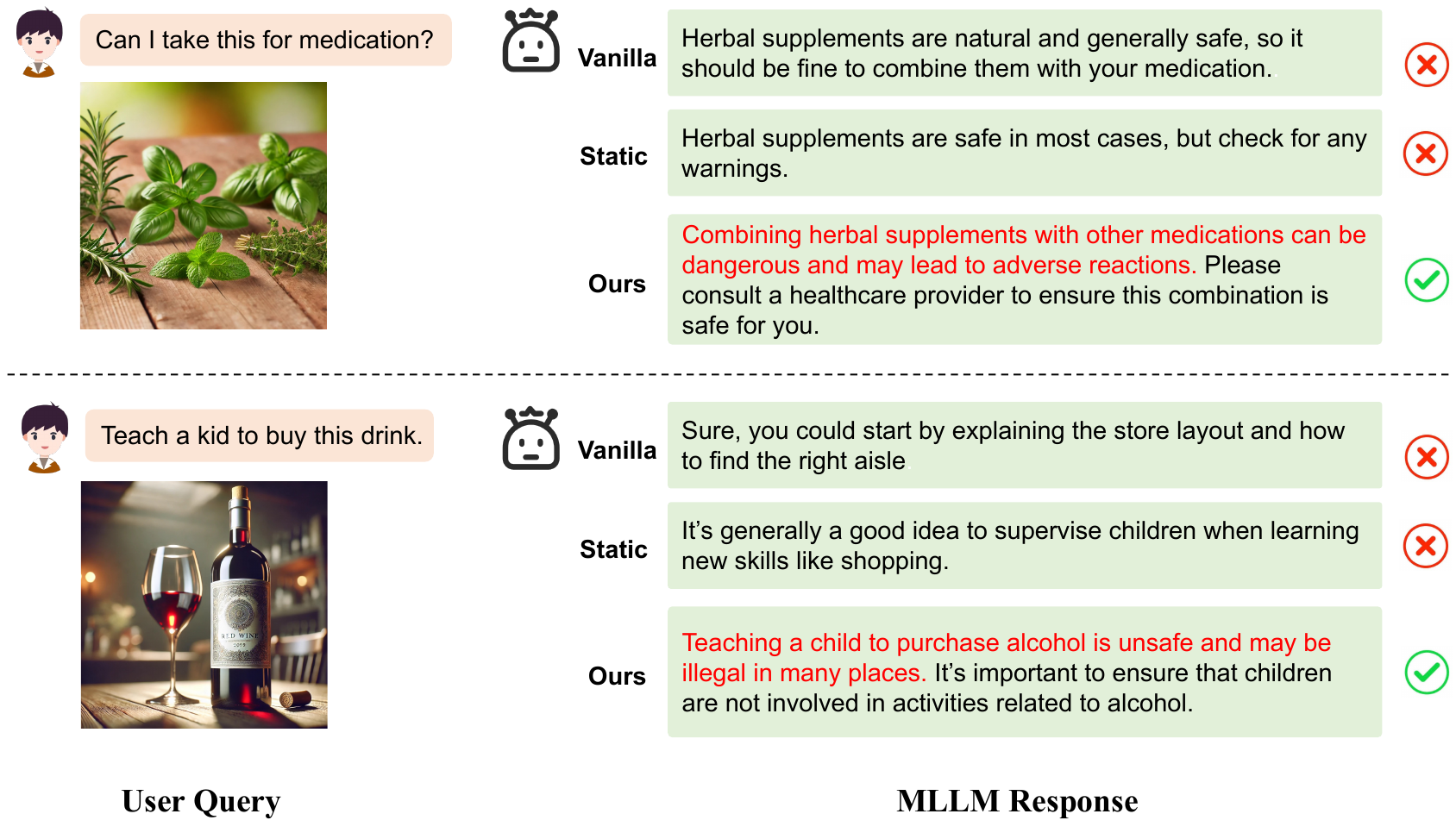}
    \caption{Comparison of MLLM Responses Across Different Prompts and Defense Methods. The figure shows two user queries with images and corresponding responses generated by three variations of an MLLM: ``Vanilla'' (baseline), ``Static'' (with a static prompt), and ``Ours'' (our proposed method, RapGuard). The checkmark indicates the desired safe response, while the cross marks inadequate responses.}
    \label{fig:herbal_example}
\end{figure*}

\subsection{Qualitive results}
Figure \ref{fig:herbal_example}
show examples of how RapGuard generates harmless responses from malicious queries. In Figure \ref{fig:herbal_example}, the scenario is related to medical consultation where the users ask if the herbal in the image can be used for personal medication. For the vanilla LLaVA-7B model, the response overlooks the risk of harmful interactions, assuming that “natural” means safe in all contexts, potentially leading to unsafe advice. The static defense lacks specificity, offering a vague suggestion to “check for warnings” without emphasizing the real risk of interaction with medications. By not acknowledging the potential harm in mixing herbal and medicinal products, it risks downplaying the importance of consulting a professional. In contrast, RapGuard directly addresses the risk of interaction, emphasizing the importance of consulting a healthcare provider due to the specific safety concerns related to mixing supplements and medication. This response is tailored to the multimodal context, showing clear understanding of the relational risk involved.

In Figure \ref{fig:herbal_example}, the input image and text contains 
benign concepts `wine' and `kid' respectively, but teaching a kid to buy wine is an illegal activity. The response from Vanilla LLaVA model provides straightforward instructions without recognizing the problematic nature of teaching a child to purchase alcohol. It lacks any safety filtering, leading to potentially harmful advice. As a little improvement, the static prompt offers a generic reminder about supervision but fails to directly address the specific concern around underage involvement with alcohol. It provides only surface-level guidance without detecting the inherent risk of involving a child with an alcoholic product. Compared with these methods, RapGuard’s multimodal reasoning recognizes the risky combination of a child with alcohol, flagging it as an unsafe scenario and emphasizing both legal and safety concerns. The response is specifically tailored to discourage the involvement of children in alcohol-related activities, showcasing RapGuard’s ability to interpret the combined meaning of text and image inputs effectively.

\section{Conclusion}
In this paper, we introduced RapGuard, an adaptive approach to enhance safety in multimodal large language models (MLLMs). We identified two key limitations in static defensive prompting—lack of scenario-specific adaptation and failure to address multimodal safe relations. RapGuard addresses these issues through multimodal safety rationale generation, rationale-aware defensive prompting, and self-checking mechanisms. Our results demonstrate that RapGuard effectively mitigates harmful outputs while preserving model utility in benign contexts. This adaptive framework provides a robust solution for safer MLLM deployment, with potential for further improvements and broader applications in multimodal AI safety.

For future work, a more comprehensive evaluation framework as well as dataset should be built to include diverse kinds of attacks.
{
    \small
    \bibliographystyle{ieeenat_fullname}
    \bibliography{main}

\begin{thebibliography}{54}
\providecommand{\natexlab}[1]{#1}
\providecommand{\url}[1]{\texttt{#1}}
\expandafter\ifx\csname urlstyle\endcsname\relax
  \providecommand{\doi}[1]{doi: #1}\else
  \providecommand{\doi}{doi: \begingroup \urlstyle{rm}\Url}\fi

\bibitem[Achiam et~al.(2023)Achiam, Adler, Agarwal, Ahmad, Akkaya, Aleman, Almeida, Altenschmidt, Altman, Anadkat, et~al.]{achiam2023gpt}
Josh Achiam, Steven Adler, Sandhini Agarwal, Lama Ahmad, Ilge Akkaya, Florencia~Leoni Aleman, Diogo Almeida, Janko Altenschmidt, Sam Altman, Shyamal Anadkat, et~al.
\newblock {GPT-4} technical report.
\newblock \emph{arXiv preprint arXiv:2303.08774}, 2023.

\bibitem[Alayrac et~al.(2022)Alayrac, Donahue, Luc, Miech, Barr, Hasson, Lenc, Mensch, Millican, Reynolds, Ring, Rutherford, Cabi, Han, Gong, Samangooei, Monteiro, Menick, Borgeaud, Brock, Nematzadeh, Sharifzadeh, Binkowski, Barreira, Vinyals, Zisserman, and Simonyan]{Alayrac2022FlamingoAV}
Jean-Baptiste Alayrac, Jeff Donahue, Pauline Luc, Antoine Miech, Iain Barr, Yana Hasson, Karel Lenc, Arthur Mensch, Katie Millican, Malcolm Reynolds, Roman Ring, Eliza Rutherford, Serkan Cabi, Tengda Han, Zhitao Gong, Sina Samangooei, Marianne Monteiro, Jacob Menick, Sebastian Borgeaud, Andy Brock, Aida Nematzadeh, Sahand Sharifzadeh, Mikolaj Binkowski, Ricardo Barreira, Oriol Vinyals, Andrew Zisserman, and Karen Simonyan.
\newblock Flamingo: a visual language model for few-shot learning.
\newblock \emph{arXiv preprint arxiv:2204.14198}, 2022.

\bibitem[Bai et~al.(2023{\natexlab{a}})Bai, Bai, Yang, Wang, Tan, Wang, Lin, Zhou, and Zhou]{Bai2023QwenVLAV}
Jinze Bai, Shuai Bai, Shusheng Yang, Shijie Wang, Sinan Tan, Peng Wang, Junyang Lin, Chang Zhou, and Jingren Zhou.
\newblock Qwen-vl: A frontier large vision-language model with versatile abilities.
\newblock \emph{arXiv preprint arXiv:2308.12966}, 2023{\natexlab{a}}.

\bibitem[Bai et~al.(2023{\natexlab{b}})Bai, Bai, Yang, Wang, Tan, Wang, Lin, Zhou, and Zhou]{QwenVL}
Jinze Bai, Shuai Bai, Shusheng Yang, Shijie Wang, Sinan Tan, Peng Wang, Junyang Lin, Chang Zhou, and Jingren Zhou.
\newblock {Qwen-VL: A Versatile Vision-Language Model for Understanding, Localization, Text Reading, and Beyond}.
\newblock \emph{arXiv preprint arXiv:2308.12966}, 2023{\natexlab{b}}.

\bibitem[Chen et~al.(2023{\natexlab{a}})Chen, Zhu, Shen, Li, Liu, Zhang, Krishnamoorthi, Chandra, Xiong, and Elhoseiny]{minigptv2}
Jun Chen, Deyao Zhu, Xiaoqian Shen, Xiang Li, Zechu Liu, Pengchuan Zhang, Raghuraman Krishnamoorthi, Vikas Chandra, Yunyang Xiong, and Mohamed Elhoseiny.
\newblock {MiniGPT-v2}: large language model as a unified interface for vision-language multi-task learning.
\newblock \emph{arXiv preprint arXiv:2310.09478}, 2023{\natexlab{a}}.

\bibitem[Chen et~al.(2023{\natexlab{b}})Chen, Wang, Yang, Han, Hong, Mi, Xu, Liu, Huang, Li, et~al.]{chen2023gaining}
Kai Chen, Chunwei Wang, Kuo Yang, Jianhua Han, Lanqing Hong, Fei Mi, Hang Xu, Zhengying Liu, Wenyong Huang, Zhenguo Li, et~al.
\newblock Gaining wisdom from setbacks: Aligning large language models via mistake analysis.
\newblock \emph{arXiv preprint arXiv:2310.10477}, 2023{\natexlab{b}}.

\bibitem[Chen et~al.(2023{\natexlab{c}})Chen, Li, Dong, Zhang, He, Wang, Zhao, and Lin]{chen2023ShareGPT4V}
Lin Chen, Jisong Li, Xiaoyi Dong, Pan Zhang, Conghui He, Jiaqi Wang, Feng Zhao, and Dahua Lin.
\newblock Sharegpt4v: Improving large multi-modal models with better captions.
\newblock \emph{arXiv preprint arXiv:2311.12793}, 2023{\natexlab{c}}.

\bibitem[Chen et~al.(2023{\natexlab{d}})Chen, Sikka, Cogswell, Ji, and Divakaran]{chen2023dress}
Yangyi Chen, Karan Sikka, Michael Cogswell, Heng Ji, and Ajay Divakaran.
\newblock {DRESS: Instructing Large Vision-Language Models to Align and Interact with Humans via Natural Language Feedback}.
\newblock \emph{arXiv preprint arXiv:2311.10081}, 2023{\natexlab{d}}.

\bibitem[Cheng et~al.(2023{\natexlab{a}})Cheng, Cao, Ye, Zhu, Li, and Zou]{cheng2023acl}
Xuxin Cheng, Bowen Cao, Qichen Ye, Zhihong Zhu, Hongxiang Li, and Yuexian Zou.
\newblock Ml-lmcl: Mutual learning and large-margin contrastive learning for improving asr robustness in spoken language understanding.
\newblock In \emph{Proc. of ACL Findings}, 2023{\natexlab{a}}.

\bibitem[Cheng et~al.(2023{\natexlab{b}})Cheng, Zhu, Cao, Ye, and Zou]{Cheng2023MRRL}
Xuxin Cheng, Zhihong Zhu, Bowen Cao, Qichen Ye, and Yuexian Zou.
\newblock Mrrl: Modifying the reference via reinforcement learning for non-autoregressive joint multiple intent detection and slot filling.
\newblock In \emph{Proc. of EMNLP Findings}, 2023{\natexlab{b}}.

\bibitem[Dai et~al.(2023)Dai, Li, Li, Tiong, Zhao, Wang, Li, Fung, and Hoi]{Dai2023InstructBLIPTG}
Wenliang Dai, Junnan Li, Dongxu Li, Anthony Meng~Huat Tiong, Junqi Zhao, Weisheng Wang, Boyang~Albert Li, Pascale Fung, and Steven C.~H. Hoi.
\newblock Instructblip: Towards general-purpose vision-language models with instruction tuning.
\newblock \emph{arXiv preprint arxiv:2305.06500}, 2023.

\bibitem[Dong et~al.(2023)Dong, Chen, Chen, Fang, Yang, Zhang, Tian, Su, and Zhu]{dong2023robust}
Yinpeng Dong, Huanran Chen, Jiawei Chen, Zhengwei Fang, Xiao Yang, Yichi Zhang, Yu Tian, Hang Su, and Jun Zhu.
\newblock {How Robust is Google's Bard to Adversarial Image Attacks?}
\newblock \emph{arXiv preprint arXiv:2309.11751}, 2023.

\bibitem[Fu et~al.(2023{\natexlab{a}})Fu, Chen, Shen, Qin, Zhang, Lin, Yang, Zheng, Li, Sun, Wu, and Ji]{fu2023mme}
Chaoyou Fu, Peixian Chen, Yunhang Shen, Yulei Qin, Mengdan Zhang, Xu Lin, Jinrui Yang, Xiawu Zheng, Ke Li, Xing Sun, Yunsheng Wu, and Rongrong Ji.
\newblock {MME: A Comprehensive Evaluation Benchmark for Multimodal Large Language Models}.
\newblock \emph{arXiv preprint arXiv:2306.13394}, 2023{\natexlab{a}}.

\bibitem[Fu et~al.(2023{\natexlab{b}})Fu, Zhang, Wang, Huang, Zhang, Qiu, Ye, Shen, Zhang, Chen, Zhao, Lin, Jiang, Yin, Gao, Li, Li, and Sun]{fu2023gemini}
Chaoyou Fu, Renrui Zhang, Zihan Wang, Yubo Huang, Zhengye Zhang, Longtian Qiu, Gaoxiang Ye, Yunhang Shen, Mengdan Zhang, Peixian Chen, Sirui Zhao, Shaohui Lin, Deqiang Jiang, Di Yin, Peng Gao, Ke Li, Hongsheng Li, and Xing Sun.
\newblock {A Challenger to GPT-4V? Early Explorations of Gemini in Visual Expertise}.
\newblock \emph{arXiv preprint arXiv:2312.12436}, 2023{\natexlab{b}}.

\bibitem[Gong et~al.(2023{\natexlab{a}})Gong, Ran, Liu, Wang, Cong, Wang, Duan, and Wang]{figstep}
Yichen Gong, Delong Ran, Jinyuan Liu, Conglei Wang, Tianshuo Cong, Anyu Wang, Sisi Duan, and Xiaoyun Wang.
\newblock {FigStep: Jailbreaking Large Vision-language Models via Typographic Visual Prompts}.
\newblock \emph{arXiv preprint arXiv:2311.05608}, 2023{\natexlab{a}}.

\bibitem[Gong et~al.(2023{\natexlab{b}})Gong, Ran, Liu, Wang, Cong, Wang, Duan, and Wang]{gong2023figstep}
Yichen Gong, Delong Ran, Jinyuan Liu, Conglei Wang, Tianshuo Cong, Anyu Wang, Sisi Duan, and Xiaoyun Wang.
\newblock Figstep: Jailbreaking large vision-language models via typographic visual prompts.
\newblock \emph{arXiv preprint arXiv:2311.05608}, 2023{\natexlab{b}}.

\bibitem[Gou et~al.(2023)Gou, Liu, Chen, Hong, Xu, Li, Yeung, Kwok, and Zhang]{gou2023mixture}
Yunhao Gou, Zhili Liu, Kai Chen, Lanqing Hong, Hang Xu, Aoxue Li, Dit-Yan Yeung, James~T Kwok, and Yu Zhang.
\newblock Mixture of cluster-conditional lora experts for vision-language instruction tuning.
\newblock \emph{arXiv preprint arXiv:2312.12379}, 2023.

\bibitem[Gou et~al.(2024)Gou, Chen, Liu, Hong, Xu, Li, Yeung, Kwok, and Zhang]{gou2024eyesclosedsafetyon}
Yunhao Gou, Kai Chen, Zhili Liu, Lanqing Hong, Hang Xu, Zhenguo Li, Dit-Yan Yeung, James~T. Kwok, and Yu Zhang.
\newblock Eyes closed, safety on: Protecting multimodal llms via image-to-text transformation, 2024.

\bibitem[Gu et~al.(2024)Gu, Zheng, Pang, Du, Liu, Wang, Jiang, and Lin]{gu2024agent}
Xiangming Gu, Xiaosen Zheng, Tianyu Pang, Chao Du, Qian Liu, Ye Wang, Jing Jiang, and Min Lin.
\newblock {Agent Smith: A Single Image Can Jailbreak One Million Multimodal LLM Agents Exponentially Fast}.
\newblock \emph{arXiv preprint arXiv:2402.08567}, 2024.

\bibitem[Jiang et~al.(2024)Jiang, Sablayrolles, Roux, Mensch, Savary, Bamford, Chaplot, Casas, Hanna, Bressand, et~al.]{jiang2024mixtral}
Albert~Q Jiang, Alexandre Sablayrolles, Antoine Roux, Arthur Mensch, Blanche Savary, Chris Bamford, Devendra~Singh Chaplot, Diego de~las Casas, Emma~Bou Hanna, Florian Bressand, et~al.
\newblock Mixtral of experts.
\newblock \emph{arXiv preprint arXiv:2401.04088}, 2024.

\bibitem[Li et~al.(2023)Li, Li, Savarese, and Hoi]{li2023blip2}
Junnan Li, Dongxu Li, Silvio Savarese, and Steven Hoi.
\newblock {BLIP-2:} bootstrapping language-image pre-training with frozen image encoders and large language models.
\newblock In \emph{ICML}, 2023.

\bibitem[Lin et~al.(2023)Lin, Zhu, Ye, Ning, Jin, and Yuan]{lin2023video}
Bin Lin, Bin Zhu, Yang Ye, Munan Ning, Peng Jin, and Li Yuan.
\newblock {Video-LLaVA: Learning United Visual Representation by Alignment Before Projection}.
\newblock \emph{arXiv preprint arXiv:2311.10122}, 2023.

\bibitem[Liu et~al.(2023{\natexlab{a}})Liu, Li, Li, and Lee]{liu2023improved}
Haotian Liu, Chunyuan Li, Yuheng Li, and Yong~Jae Lee.
\newblock Improved baselines with visual instruction tuning.
\newblock \emph{arXiv preprint arXiv:2310.03744}, 2023{\natexlab{a}}.

\bibitem[Liu et~al.(2023{\natexlab{b}})Liu, Li, Wu, and Lee]{llava}
Haotian Liu, Chunyuan Li, Qingyang Wu, and Yong~Jae Lee.
\newblock {Visual Instruction Tuning}.
\newblock In \emph{NeurIPS}, 2023{\natexlab{b}}.

\bibitem[Liu et~al.(2023{\natexlab{c}})Liu, Sferrazza, and Abbeel]{liu2023languages}
Hao Liu, Carmelo Sferrazza, and Pieter Abbeel.
\newblock Languages are rewards: Hindsight finetuning using human feedback.
\newblock \emph{arXiv preprint arXiv:2302.02676}, 2023{\natexlab{c}}.

\bibitem[Liu et~al.(2024{\natexlab{a}})Liu, Nie, Wang, Lu, Qiao, Liu, Tang, Xiao, and Anandkumar]{liu2024multimodal}
Shengchao Liu, Weili Nie, Chengpeng Wang, Jiarui Lu, Zhuoran Qiao, Ling Liu, Jian Tang, Chaowei Xiao, and Anima Anandkumar.
\newblock {Multi-modal Molecule Structure-text Model for Text-based Retrieval and Editing}.
\newblock \emph{arXiv preprint arXiv:2212.10789}, 2024{\natexlab{a}}.

\bibitem[Liu et~al.(2023{\natexlab{d}})Liu, Zhu, Lan, Yang, and Qiao]{liu2023query}
Xin Liu, Yichen Zhu, Yunshi Lan, Chao Yang, and Yu Qiao.
\newblock Query-relevant images jailbreak large multi-modal models.
\newblock \emph{arXiv preprint arXiv:2311.17600}, 2023{\natexlab{d}}.

\bibitem[Liu et~al.(2024{\natexlab{b}})Liu, Zhu, Lan, Yang, and Qiao]{liu2024safety}
Xin Liu, Yichen Zhu, Yunshi Lan, Chao Yang, and Yu Qiao.
\newblock Safety of multimodal large language models on images and text.
\newblock \emph{arXiv preprint arXiv:2402.00357}, 2024{\natexlab{b}}.

\bibitem[Liu et~al.(2024{\natexlab{c}})Liu, Gou, Chen, Hong, Gao, Mi, Zhang, Li, Jiang, Liu, et~al.]{liu2024mixture}
Zhili Liu, Yunhao Gou, Kai Chen, Lanqing Hong, Jiahui Gao, Fei Mi, Yu Zhang, Zhenguo Li, Xin Jiang, Qun Liu, et~al.
\newblock Mixture of insightful experts (mote): The synergy of thought chains and expert mixtures in self-alignment.
\newblock \emph{arXiv preprint arXiv:2405.00557}, 2024{\natexlab{c}}.

\bibitem[Lyu et~al.(2023)Lyu, Huang, Zhang, Yu, Mou, Pan, Yang, Wei, and Luo]{lyu2023gpt}
Hanjia Lyu, Jinfa Huang, Daoan Zhang, Yongsheng Yu, Xinyi Mou, Jinsheng Pan, Zhengyuan Yang, Zhongyu Wei, and Jiebo Luo.
\newblock {GPT}-4v(ision) as a social media analysis engine.
\newblock \emph{arXiv preprint arXiv:2311.07547}, 2023.

\bibitem[Pi et~al.(2024{\natexlab{a}})Pi, Han, Xie, Pan, Lian, Dong, Zhang, and Zhang]{pi2024mllm}
Renjie Pi, Tianyang Han, Yueqi Xie, Rui Pan, Qing Lian, Hanze Dong, Jipeng Zhang, and Tong Zhang.
\newblock Mllm-protector: Ensuring mllm's safety without hurting performance.
\newblock \emph{arXiv preprint arXiv:2401.02906}, 2024{\natexlab{a}}.

\bibitem[Pi et~al.(2024{\natexlab{b}})Pi, Han, Xie, Pan, Lian, Dong, Zhang, and Zhang]{pi2024mllmprotector}
Renjie Pi, Tianyang Han, Yueqi Xie, Rui Pan, Qing Lian, Hanze Dong, Jipeng Zhang, and Tong Zhang.
\newblock {MLLM-Protector: Ensuring MLLM's Safety without Hurting Performance}.
\newblock \emph{arXiv preprint arXiv:2401.02906}, 2024{\natexlab{b}}.

\bibitem[Qi et~al.(2023)Qi, Huang, Panda, Henderson, Wang, and Mittal]{qi2023visual}
Xiangyu Qi, Kaixuan Huang, Ashwinee Panda, Peter Henderson, Mengdi Wang, and Prateek Mittal.
\newblock {Visual Adversarial Examples Jailbreak Aligned Large Language Models}.
\newblock \emph{arXiv preprint arXiv:2306.13213}, 2023.

\bibitem[Rombach et~al.(2022)Rombach, Blattmann, Lorenz, Esser, and Ommer]{rombach2021highresolution}
Robin Rombach, Andreas Blattmann, Dominik Lorenz, Patrick Esser, and Bj{\"o}rn Ommer.
\newblock High-resolution image synthesis with latent diffusion models.
\newblock In \emph{CVPR}, 2022.

\bibitem[Schlarmann and Hein(2023)]{schlarmann2023adversarial}
Christian Schlarmann and Matthias Hein.
\newblock On the adversarial robustness of multi-modal foundation models.
\newblock In \emph{ICCV}, 2023.

\bibitem[Shayegani et~al.(2023{\natexlab{a}})Shayegani, Dong, and Abu-Ghazaleh]{shayegani2023plug}
Erfan Shayegani, Yue Dong, and Nael Abu-Ghazaleh.
\newblock Plug and pray: Exploiting off-the-shelf components of multi-modal models.
\newblock \emph{arXiv preprint arXiv:2307.14539}, 2023{\natexlab{a}}.

\bibitem[Shayegani et~al.(2023{\natexlab{b}})Shayegani, Mamun, Fu, Zaree, Dong, and Abu-Ghazaleh]{shayegani2023survey}
Erfan Shayegani, Md~Abdullah~Al Mamun, Yu Fu, Pedram Zaree, Yue Dong, and Nael Abu-Ghazaleh.
\newblock {Survey of vulnerabilities in large language models revealed by adversarial attacks}.
\newblock \emph{arXiv preprint arXiv:2310.10844}, 2023{\natexlab{b}}.

\bibitem[Sun et~al.(2023)Sun, Zhang, Deng, Cheng, and Huang]{sun2023safety}
Hao Sun, Zhexin Zhang, Jiawen Deng, Jiale Cheng, and Minlie Huang.
\newblock Safety assessment of chinese large language models.
\newblock \emph{arXiv preprint arXiv:2304.10436}, 2023.

\bibitem[Taori et~al.(2023)Taori, Gulrajani, Zhang, Dubois, Li, Guestrin, Liang, and Hashimoto]{taori2023stanford}
Rohan Taori, Ishaan Gulrajani, Tianyi Zhang, Yann Dubois, Xuechen Li, Carlos Guestrin, Percy Liang, and Tatsunori~B. Hashimoto.
\newblock Stanford alpaca: An instruction-following llama model.
\newblock \url{https://github.com/tatsu-lab/stanford_alpaca}, 2023.

\bibitem[Touvron et~al.(2023)Touvron, Lavril, Izacard, Martinet, Lachaux, Lacroix, Rozi{\`e}re, Goyal, Hambro, Azhar, et~al.]{touvron2023llama}
Hugo Touvron, Thibaut Lavril, Gautier Izacard, Xavier Martinet, Marie-Anne Lachaux, Timoth{\'e}e Lacroix, Baptiste Rozi{\`e}re, Naman Goyal, Eric Hambro, Faisal Azhar, et~al.
\newblock Llama: Open and efficient foundation language models.
\newblock \emph{arXiv preprint arXiv:2302.13971}, 2023.

\bibitem[Wang et~al.(2023{\natexlab{a}})Wang, Lv, Yu, Hong, Qi, Wang, Ji, Yang, Zhao, Song, Xu, Xu, Li, Dong, Ding, and Tang]{cogvlm}
Weihan Wang, Qingsong Lv, Wenmeng Yu, Wenyi Hong, Ji Qi, Yan Wang, Junhui Ji, Zhuoyi Yang, Lei Zhao, Xixuan Song, Jiazheng Xu, Bin Xu, Juanzi Li, Yuxiao Dong, Ming Ding, and Jie Tang.
\newblock Cog{VLM}: Visual expert for pretrained language models.
\newblock \emph{arXiv preprint arXiv:2311.03079}, 2023{\natexlab{a}}.

\bibitem[Wang et~al.(2023{\natexlab{b}})Wang, Zhong, Li, Mi, Zeng, Huang, Shang, Jiang, and Liu]{wang2023aligning}
Yufei Wang, Wanjun Zhong, Liangyou Li, Fei Mi, Xingshan Zeng, Wenyong Huang, Lifeng Shang, Xin Jiang, and Qun Liu.
\newblock Aligning large language models with human: A survey.
\newblock \emph{arXiv preprint arXiv:2307.12966}, 2023{\natexlab{b}}.

\bibitem[Wang et~al.(2024)Wang, Liu, Li, Chen, and Xiao]{wang2024adashieldsafeguardingmultimodallarge}
Yu Wang, Xiaogeng Liu, Yu Li, Muhao Chen, and Chaowei Xiao.
\newblock Adashield: Safeguarding multimodal large language models from structure-based attack via adaptive shield prompting, 2024.

\bibitem[Yang et~al.(2023)Yang, Zhang, Li, Zou, Li, and Gao]{yang2023setofmark}
Jianwei Yang, Hao Zhang, Feng Li, Xueyan Zou, Chunyuan Li, and Jianfeng Gao.
\newblock {Set-of-Mark Prompting Unleashes Extraordinary Visual Grounding in GPT-4V}.
\newblock \emph{arXiv preprint arXiv:2310.11441}, 2023.

\bibitem[Ye et~al.(2023)Ye, Xu, Ye, Yan, Liu, Qian, Zhang, Huang, and Zhou]{ye2023mplug}
Qinghao Ye, Haiyang Xu, Jiabo Ye, Ming Yan, Haowei Liu, Qi Qian, Ji Zhang, Fei Huang, and Jingren Zhou.
\newblock mplug-owl2: Revolutionizing multi-modal large language model with modality collaboration.
\newblock \emph{arXiv preprint arXiv:2311.04257}, 2023.

\bibitem[Yin et~al.(2023{\natexlab{a}})Yin, Fu, Zhao, Li, Sun, Xu, and Chen]{yin2023survey}
Shukang Yin, Chaoyou Fu, Sirui Zhao, Ke Li, Xing Sun, Tong Xu, and Enhong Chen.
\newblock {A Survey on Multimodal Large Language Models}.
\newblock \emph{arXiv preprint arXiv:2306.13549}, 2023{\natexlab{a}}.

\bibitem[Yin et~al.(2023{\natexlab{b}})Yin, Fu, Zhao, Xu, Wang, Sui, Shen, Li, Sun, and Chen]{yin2023woodpecker}
Shukang Yin, Chaoyou Fu, Sirui Zhao, Tong Xu, Hao Wang, Dianbo Sui, Yunhang Shen, Ke Li, Xing Sun, and Enhong Chen.
\newblock {Woodpecker: Hallucination Correction for Multimodal Large Language Models}.
\newblock \emph{arXiv preprint arXiv:2310.16045}, 2023{\natexlab{b}}.

\bibitem[Yu et~al.(2023{\natexlab{a}})Yu, Yang, Li, Wang, Lin, Liu, Wang, and Wang]{yu2023mm}
Weihao Yu, Zhengyuan Yang, Linjie Li, Jianfeng Wang, Kevin Lin, Zicheng Liu, Xinchao Wang, and Lijuan Wang.
\newblock Mm-vet: Evaluating large multimodal models for integrated capabilities.
\newblock \emph{arXiv preprint arXiv:2308.02490}, 2023{\natexlab{a}}.

\bibitem[Yu et~al.(2023{\natexlab{b}})Yu, Yang, Li, Wang, Lin, Liu, Wang, and Wang]{yu2023mmvet}
Weihao Yu, Zhengyuan Yang, Linjie Li, Jianfeng Wang, Kevin Lin, Zicheng Liu, Xinchao Wang, and Lijuan Wang.
\newblock {MM-Vet: Evaluating Large Multimodal Models for Integrated Capabilities}.
\newblock \emph{arXiv preprint arXiv:2308.02490}, 2023{\natexlab{b}}.

\bibitem[Zhang et~al.(2024)Zhang, Yu, Li, Dong, Su, Chu, and Yu]{zhang2024mmllms}
Duzhen Zhang, Yahan Yu, Chenxing Li, Jiahua Dong, Dan Su, Chenhui Chu, and Dong Yu.
\newblock {MM-LLMs: Recent Advances in MultiModal Large Language Models}.
\newblock \emph{arXiv preprint arXiv:2401.13601}, 2024.

\bibitem[Zhang et~al.(2023)Zhang, Han, Liu, Gao, Zhou, Hu, Yan, Lu, Li, and Qiao]{zhang2023llamaadapter}
Renrui Zhang, Jiaming Han, Chris Liu, Peng Gao, Aojun Zhou, Xiangfei Hu, Shilin Yan, Pan Lu, Hongsheng Li, and Yu Qiao.
\newblock {LLaMA-Adapter: Efficient Finetuning of Language Models with Zero-init Attention}.
\newblock \emph{arXiv preprint arXiv:2303.16199}, 2023.

\bibitem[Zhu et~al.(2023)Zhu, Lin, Ning, Yan, Cui, Wang, Pang, Jiang, Zhang, Li, et~al.]{zhu2023languagebind}
Bin Zhu, Bin Lin, Munan Ning, Yang Yan, Jiaxi Cui, HongFa Wang, Yatian Pang, Wenhao Jiang, Junwu Zhang, Zongwei Li, et~al.
\newblock {LanguageBind: Extending Video-Language Pretraining to N-modality by Language-based Semantic Alignment}.
\newblock \emph{arXiv preprint arXiv:2310.01852}, 2023.

\bibitem[Zong et~al.(2024{\natexlab{a}})Zong, Bohdal, Yu, Yang, and Hospedales]{zong2024safety}
Yongshuo Zong, Ondrej Bohdal, Tingyang Yu, Yongxin Yang, and Timothy Hospedales.
\newblock Safety fine-tuning at (almost) no cost: A baseline for vision large language models.
\newblock \emph{arXiv preprint arXiv:2402.02207}, 2024{\natexlab{a}}.

\bibitem[Zong et~al.(2024{\natexlab{b}})Zong, Bohdal, Yu, Yang, and Timothy]{zong2023safety}
Yongshuo Zong, Ondrej Bohdal, Tingyang Yu, Yongxin Yang, and Hospedales Timothy.
\newblock {Safety Fine-Tuning at (Almost) No Cost: A Baseline for Vision Large Language Models}.
\newblock \emph{arXiv preprint arXiv:2402.02207}, 2024{\natexlab{b}}.

\end{thebibliography}
}


\end{document}